\documentclass[conference]{IEEEtran}
\IEEEoverridecommandlockouts

\usepackage{pgfplots}
\pgfplotsset{compat=1.18}
\usepackage{tikz}
\usetikzlibrary{positioning, shapes.geometric, arrows.meta}
\usepackage{placeins}
\usepackage{float}
\usepackage{tabularx}
\usepackage{makecell}
\usepackage{booktabs} 

\usepackage{cite}
\usepackage{amsmath,amssymb,amsfonts}
\usepackage{algorithmic}
\usepackage{graphicx}
\usepackage{textcomp}
\usepackage{xcolor}
\def\BibTeX{{\rm B\kern-.05em{\sc i\kern-.025em b}\kern-.08em
    T\kern-.1667em\lower.7ex\hbox{E}\kern-.125emX}}
\begin{document}

\title{Explainable Gait Abnormality Detection Using Dual-Dataset CNN-LSTM Models}

\author{
\IEEEauthorblockN{Parth Agarwal\textsuperscript{*}, Sangaa Chatterjee\textsuperscript{*}}
\IEEEauthorblockA{\textit{Computer Science and Engineering} \\
\textit{Pennsylvania State University}\\
University Park, PA, USA \\
\{pxa5191, sjc6940\}@psu.edu}
\and
\IEEEauthorblockN{Md Faisal Kabir}
\IEEEauthorblockA{\textit{Computer Science and Engineering} \\
\textit{Pennsylvania State University}\\
Harrisburg, PA, USA \\
mpk5904@psu.edu}
\and
\IEEEauthorblockN{Suman Saha}
\IEEEauthorblockA{\textit{Computer Science and Engineering} \\
\textit{Pennsylvania State University}\\
University Park, PA, USA \\
szs339@psu.edu}

\thanks{\textsuperscript{*}Parth Agarwal and Sangaa Chatterjee contributed equally to this work.}
}

\maketitle

\begin{abstract}

Gait is a key indicator in diagnosing movement disorders, but most models lack interpretability and rely on single datasets. We propose a dual-branch CNN–LSTM framework: a 1D branch on joint-based features from GAVD and a 3D branch on silhouettes from OU-MVLP. Interpretability is provided by SHAP (temporal attributions) and Grad-CAM (spatial localization).On held-out sets, the system achieves 98.6\% accuracy with strong recall and F1.This approach advances explainable gait analysis across both clinical and biometric domains.

\end{abstract}

\begin{IEEEkeywords}
Deep learning, Gait analysis, Explainable AI, Convolutional neural networks, Long short-term memory
\end{IEEEkeywords}

\section{Introduction}

Gait analysis is widely used to diagnose and monitor movement disorders such as cerebral palsy, Parkinson’s disease, and stroke \cite{prakash2016recent}. Changes in a person's walking pattern often reveal early signs of these conditions, making gait a critical indicator in clinical settings \cite{prakash2016recent}. Traditionally, assessments have relied on human observation and expert judgment, which can vary between practitioners and may not scale well to large patient populations \cite{prakash2016recent}.With the rise of machine learning, deep learning models have shown strong potential for automating gait analysis \cite{shen2022comprehensive, munusamy2024emerging}.

Yet most existing models lack interpretability, a major barrier for clinical adoption since clinicians must understand why a decision was made \cite{chen2020explainable, sljepcevic2023explainable, ozates2024identification}. Many approaches also focus on silhouettes for biometric identification \cite{gaitgl2021, armin2018hybrid}, overlooking clinical cues such as joint motion and step timing\cite{rahman2021cnn}. In addition, reliance on a single dataset reduces generalizability: clinical datasets are often small and imbalanced, while large biometric datasets lack medically relevant detail \cite{ranjan2019gait}, \cite{takemura2020ou}. To address these gaps, we propose a dual-branch deep learning framework that combines two complementary architectures:

\begin{itemize}
    \item A 1D CNN-LSTM model for analyzing joint-based features from the Gait Abnormality Video Dataset (GAVD) \cite{ranjan2019gait}, designed for clinical diagnosis.

    \item A 3D CNN-LSTM model that processes silhouette sequences from the OU-ISIR MVLP dataset \cite{takemura2020ou} to support large-scale gait classification.
    
\end{itemize}

In addition to predictive performance, our system integrates SHAP for temporal attribution and Grad-CAM for spatial localization, offering transparency into the features driving predictions \cite{sljepcevic2023explainable, ozates2024identification, kim2022explainable}. This dual-dataset, interpretable framework addresses both clinical relevance and generalization, two central challenges in modern gait analysis.

The remainder of the paper covers related work (Section II), our framework and datasets (Section III), results and discussion (Section IV), and limitations with future directions (Section V)

\section{Related Work}

Gait analysis has become increasingly important for diagnosing neurological and musculoskeletal conditions. Traditional lab-based systems like motion capture or pressure-sensitive walkways provide precise measurements but are often expensive, time-consuming, and impractical for routine screening or home-based monitoring \cite{prakash2016recent}. To address these limitations, researchers have shifted toward video-based gait analysis powered by machine learning and computer vision \cite{shen2022comprehensive, munusamy2024emerging}. A number of large datasets have facilitated progress in this area. The Gait Abnormality Video Dataset (GAVD) provides over 1,800 annotated gait clips for clinical use \cite{ranjan2019gait}, while OU-MVLP captures gait from over 10,000 subjects across 14 viewpoints, making it a valuable benchmark for cross-view and biometric applications \cite{takemura2020ou}.

Many gait recognition systems rely on CNN-LSTM architectures to model spatial and temporal dynamics. CNNs extract visual features from each frame, while LSTMs capture sequential patterns such as stride regularity or limb coordination. Armin et al. proposed one of the early hybrid frameworks using this approach \cite{armin2018hybrid}, followed by Rahman et al., who demonstrated real-time anomaly detection using a 1D CNN-LSTM pipeline \cite{rahman2021cnn}. Other works have shown CNN-LSTM's robustness to occlusion and background clutter, making it a practical choice in clinical environments \cite{sayeed2022deep}, \cite{burges2024gait}, \cite{kumar2024optimized}.

To improve spatial representation, recent models, such as GaitGL, introduce graph-based representations to encode body structure more effectively \cite{gaitgl2021}, while GaitFormer utilizes Transformer-based temporal modeling for capturing long-range gait dependencies \cite{gaitformer2021transformer}. Xiao et al. further applied Graph Convolutional Networks (GCNs) to assess lower-limb rehabilitation, showing promise for clinical assessment beyond identification tasks \cite{xiao2024research}. Other variants include DeepWalk-initialized CNN-LSTM pipelines \cite{tiwari2025gait} and Siamese recurrent models that learn gait similarity from landmark sequences \cite{progga2024bidirectional}.

While these approaches demonstrate significant advancements, they often overlook interpretability, which is crucial in healthcare applications. Models like GaitFormer and GaitGL excel at capturing long-range temporal and spatial dependencies but lack explicit mechanisms to explain their decisions, which limits their trustworthiness in clinical settings. Furthermore, the reliance on large, often homogeneous datasets in models like DeepWalk and GCN-based frameworks poses challenges in generalization to smaller, more diverse clinical datasets. In contrast, our model incorporates a dual-dataset approach, combining both clinical (GAVD) and biometric (OU-MVLP) datasets, and integrates explainability tools such as SHAP and Grad-CAM to provide temporal and spatial insights, which directly address the black-box nature of deep learning models.

Recent works have also started to explore the integration of explainable AI (XAI) in gait analysis. Chen et al. \cite{chen2020explainable} and Sljepcevic et al. \cite{sljepcevic2023explainable} demonstrated the use of XAI techniques to enhance model transparency in gait recognition, but their approaches often focus on static features or limited temporal data. Our model extends this by leveraging both SHAP for joint-based feature importance and Grad-CAM for spatial localization, providing a comprehensive explanation of gait anomalies over time and across different body regions. Additionally, we employ SMOTE to address class imbalance, ensuring that our model is robust even in clinical settings where abnormal gait types are underrepresented.

Broad surveys by Shen et al. and Prakash et al. summarize these developments, emphasizing the growing intersection between gait recognition and healthcare applications \cite{shen2022comprehensive}, \cite{prakash2016recent}. These works stress the importance of integrating interpretable machine learning with large and diverse datasets, an approach we adopt by combining CNN-LSTM models with explainability tools across both clinical and biometric datasets. Our model not only performs at the state-of-the-art level in terms of accuracy and F1-score but also provides critical interpretability features that make it suitable for real-world clinical deployment.

\section{System Overview}

In this section, we walk through the design and implementation of our gait classification system. We start by explaining how the data was prepared, including the specific steps taken for each dataset. We then describe the structure of the model, which combines convolutional and recurrent layers to handle both spatial patterns and temporal motion. Finally, we outline the training setup and the methods we used to make the model’s decisions more understandable.

\subsection{Dataset}

We utilize two publicly available datasets in this study, \textbf{GAVD} and \textbf{OU-MVLP}, which enable us to cover both clinical and biometric perspectives of gait analysis.

The GAVD dataset contains 1,791 RGB video clips of people walking, each showing a full gait cycle. 
Each video is labeled as either normal or one of four abnormal gait types: antalgic, lurch, spastic, or steppage. These classifications are valuable for clinical diagnosis and monitoring patient conditions.

The OU-MVLP dataset, on the other hand, comprises silhouette sequences of over 10,000 individuals captured from 14 different camera angles. Although it doesn’t have medical labels, the large number of subjects and viewing angles make it valuable for testing gait models that need to work across different perspectives.

As summarized in Table \ref{tab:datasets}, the two datasets differ in several ways, including the number of people they include, the type of data used (RGB vs. silhouette), and the types of labels available. GAVD is more clinical, while OU-MVLP is better suited for biometric tasks. Figure \ref{fig:gavd_oumvlp_qualitative} shows sample frames from each dataset to provide a better sense of their visual differences.

\begin{table*}[h!]
\centering
\caption{Key Characteristics of Datasets Used}
\resizebox{\textwidth}{!}{
\begin{tabular}{|c|c|c|}
\hline
\textbf{Feature} & \textbf{GAVD} & \textbf{OU-MVLP} \\
\hline
\textbf{Number of Subjects} & 1,800 & 10,000 \\
\hline
\textbf{Type of Data} & Video Sequences & Silhouette Sequences \\
\hline
\textbf{Data Collection Method} & Video Recordings & Motion Capture \\
\hline
\textbf{Number of Views} & Single View & 14 Views \\
\hline
\textbf{Activity Types} & Normal, Abnormal & Normal \\
\hline
\textbf{Resolution} & Varies per video & 464x464 px \\
\hline
\textbf{Preprocessing} & Feature Extraction & Normalization, Standardization \\
\hline
\textbf{Data Augmentation} & Temporal Shifting, Gaussian Noise & Temporal Shifting, Gaussian Noise \\
\hline
\end{tabular}
}
\label{tab:datasets}
\end{table*}

\subsection{Preprocessing and Augmentation}

Because the two datasets differ in format and task relevance, we applied tailored preprocessing. For GAVD, OpenPose extracted 2D coordinates (x, y) for 18 joints per frame, flattened into 36-D vectors. Low-variance joints were removed using a variance threshold to reduce dimensionality and noise. To improve generalizability, light augmentation (Gaussian noise and temporal shifts) simulated natural gait variability. Given the dataset’s imbalance across abnormal gait types, we applied SMOTE (Synthetic Minority Over-sampling Technique) on the training set to generate synthetic samples of minority classes, ensuring a more balanced distribution without discarding original data.

For OU-MVLP, binary silhouette images were resized to 128×88 pixels and standardized to fixed-length sequences. For 3D CNN input, silhouettes were further resized to 44×64 and sequences padded or cropped to 50 frames, producing inputs of shape 50×44×64 optimized for spatio-temporal learning in the CNN–LSTM pipeline.

\begin{figure}[h]
\centering
\includegraphics[width=0.47\textwidth]{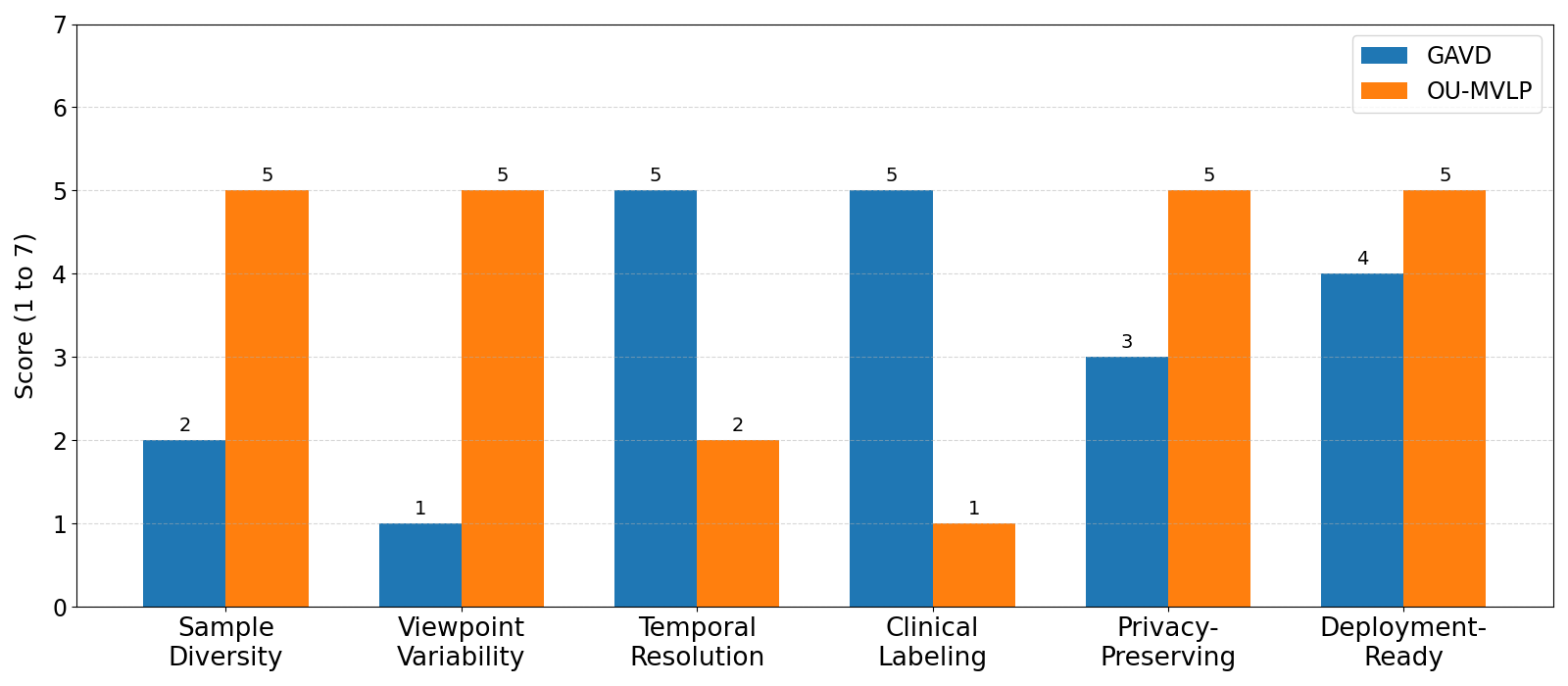}
\caption{Qualitative comparison of GAVD and OU-MVLP datasets across six dimensions. GAVD offers temporal and clinical richness; OU-MVLP supports large-scale, privacy-safe modeling.}
\label{fig:gavd_oumvlp_qualitative}
\end{figure}

Both datasets used an 80:20 subject-independent train–test split.Figure \ref{fig:gavd_oumvlp_qualitative} contrasts the RGB joint inputs from GAVD with the binary silhouettes from OU-MVLP, while Table I summarizes dataset characteristics and preprocessing details.

\subsection{Model Architecture}

The proposed model is a hybrid CNN–LSTM architecture that integrates both spatial and temporal feature extraction for robust gait classification. As shown in Figure 2, the model comprises two processing branches for handling joint-based and silhouette-based data streams. The CNN component includes three stacked 1D convolutional layers with filter sizes of 128, 256, and 512, each followed by batch normalization, ReLU activation, max pooling, and dropout for regularization. After spatial features are extracted, the output is reshaped and passed through two LSTM layers with 256 and 128 hidden units to capture the temporal dynamics of gait sequences. This is followed by a fully connected dense layer with 256 neurons and ReLU activation. The final output layer is dataset-specific: a sigmoid layer is used for binary classification in GAVD, while a softmax layer enables multiclass identity recognition in OU-MVLP. This architecture offers a balance between expressive power and parameter efficiency, making it suitable for both clinical and biometric gait analysis in practical settings.

\subsection{Integration of Explainability Tools}
To enhance model transparency and build trust in classification decisions, we integrated two complementary post hoc explainability tools: Grad-CAM and SHAP. Grad-CAM (Gradient-weighted Class Activation Mapping) is used to generate heatmaps over silhouette frames, identifying spatial regions that contributed most to the model’s decision. This is particularly useful in clinical settings, where understanding which body regions triggered a “gait anomaly” prediction can offer actionable insight for physiotherapists or neurologists.

SHAP (SHapley Additive exPlanations), on the other hand, provides frame-level importance values across the temporal dimension of each gait sequence. It quantifies how much each timestep (i.e., silhouette frame) influenced the final classification output, offering a sequence-level attribution profile. This is critical for understanding not just where in the frame, but when in the gait cycle the model detected abnormality.

Grad-CAM was chosen for its visual clarity and ability to localize spatial discriminative regions within convolutional layers—ideal for interpreting silhouette-based input. SHAP was selected due to its strong theoretical foundation in game theory and its ability to provide additive, consistent explanations across sequential inputs, making it well-suited for time-series gait data. Together, these tools address both spatial and temporal interpretability, helping bridge the gap between black-box deep models and clinical decision-making.

\begin{table}[H]
\centering
\renewcommand{\arraystretch}{1} 
\footnotesize 
\setlength{\tabcolsep}{1pt} 
\begin{tabular}{|p{1.18cm}|p{1.48cm}|p{1.38cm}|p{1.23cm}|p{1.45cm}|}
\hline
\makecell{\textbf{Dataset}} & 
\makecell{\textbf{Prep.}} & 
\makecell{\textbf{CNN}\\\textbf{Stack}} & 
\makecell{\textbf{LSTM}} & 
\makecell{\textbf{Output}\\\textbf{Layer}} \\
\hline
GAVD & 
\makecell{Joint\\features,\\norm.} & 
\makecell{Conv1\\Conv2\\Conv3} & 
\makecell{Temporal\\modeling} & 
\makecell{Normal/\\Abnormal\\classif.} \\
\hline
OU-MVLP & 
\makecell{Silhouette\\norm.} & 
\multicolumn{2}{|p{2.61cm}|}{\centering\makecell{Same CNN and LSTM\\(shared arch.)}} & 
\makecell{Identity\\classif.} \\
\hline
\end{tabular}
\caption{Overall pipeline of the proposed gait classification framework.}
\label{tab:pipeline-table}
\end{table}

\section{Experimental Setup and Results}

This section presents our experimental framework and key findings. We begin by describing the implementation details, training configuration, and evaluation metrics used to assess our model. Performance is then reported across both clinical (GAVD) and biometric (OU-MVLP) datasets, highlighting the effectiveness of our dual-path architecture. To better understand the contribution of each design choice, we perform an ablation study, evaluating model variants with different architectural components. 
Lastly, we discuss the results of our explainability modules, Grad-CAM and SHAP, which provide both visual and temporal insights into how the model makes decisions.

\subsection{Training Setup}

We implemented two tailored deep learning pipelines to address the binary classification task across two gait datasets, GAVD and OU-MVLP, each with distinct spatial-temporal characteristics.

For the GAVD dataset, we employed a CNN-LSTM architecture where each silhouette frame was flattened and passed through temporal convolutions followed by bidirectional LSTM layers. The model was trained using the Adam optimizer with a learning rate of 0.001. A batch size of 64 was chosen based on preliminary experiments showing the best balance between training speed and model performance. Dropout layers were used at a rate of 0.5 to mitigate overfitting. Hyperparameter optimization was carried out using random search over a predefined grid of learning rates (ranging from 1e-4 to 1e-2), batch sizes (32, 64, 128), and dropout rates (0.3, 0.5, 0.7). The best-performing combination was selected based on validation performance.

For the OU-MVLP dataset, we developed a 3D CNN + LSTM pipeline that preserved spatial and temporal structure. The silhouette sequences were preprocessed to (50, 44, 64) and normalized before being passed through 3D convolutional layers, followed by bidirectional LSTM and fully connected layers. We used the Adam optimizer (learning rate = 1e-4) with BCEWithLogits loss and class weights for handling class imbalance. Early stopping was applied with a patience of 10 epochs. The same random search method was applied to optimize the learning rate and batch size for this pipeline.

Both models were trained for up to 100 epochs with early stopping based on validation loss. Our results showed stable convergence with minimal overfitting, while explainability tools offered insights into the learned representations across both datasets.

\begin{figure}[htbp]
\centering
\includegraphics[width=0.45\textwidth]{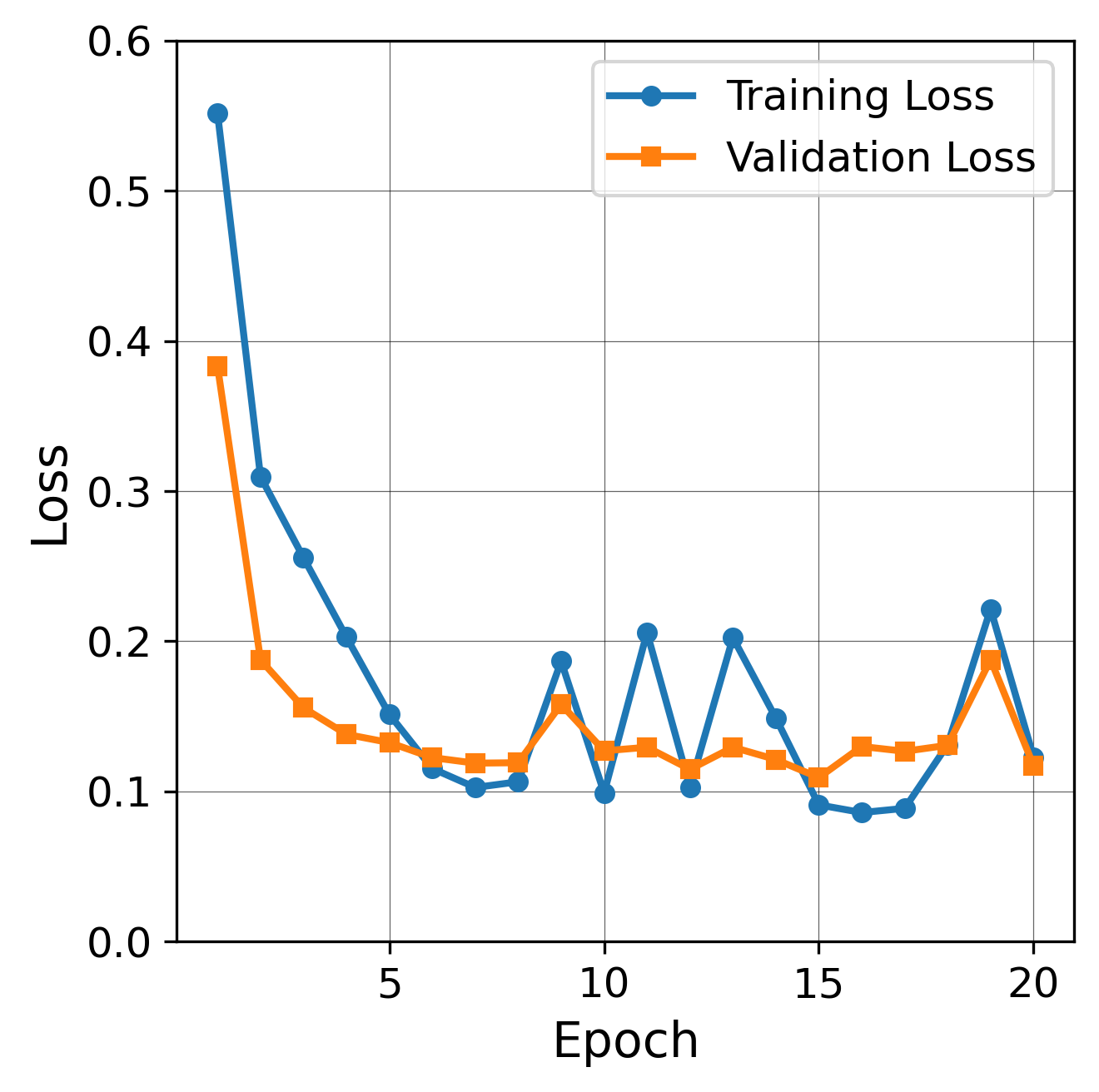}
\caption{Training and validation loss over epochs. The model exhibits effective convergence with early stopping limiting overfitting.}
\label{fig:loss-curve}
\end{figure}

Figure \ref{fig:loss-curve} shows the training and validation loss curves over epochs, demonstrating stable convergence. The training loss steadily decreases, while the validation loss flattens early, indicating effective generalization. The use of early stopping and dropout helped prevent overfitting. The small gap between the two curves suggests the model is well-regularized and performs consistently across both datasets.








\subsection{Model Performance}

We evaluated the dual-path CNN–LSTM on held-out test sets from both the GAVD and OU-MVLP datasets. Performance was measured using accuracy, precision, recall, and F1-score with the Scikit-learn evaluation suite. Accuracy provides a broad correctness measure, but since GAVD is imbalanced across abnormal gait types, we emphasized precision, recall, and F1-score to better capture diagnostic reliability. The model achieved consistently high scores across both datasets, confirming its effectiveness in learning spatial–temporal features from diverse inputs.

\begin{figure}[htbp]
\centering
\begin{tikzpicture}
\begin{axis}[
    xlabel={Epoch},
    ylabel={Accuracy},
    title={Training and Validation Accuracy over Epochs},
    legend pos=south east,
    grid=major,
    ymin=0.9, ymax=1.0
]
\addplot[color=blue, mark=*] coordinates {
  (1,0.92) (2,0.94) (3,0.96) (4,0.97) (5,0.976) (6,0.983)
};
\addlegendentry{Training}

\addplot[color=red, mark=triangle*] coordinates {
  (1,0.91) (2,0.93) (3,0.95) (4,0.965) (5,0.972) (6,0.986)
};
\addlegendentry{Validation}
\end{axis}
\end{tikzpicture}
\caption{Training and validation accuracy across epochs, showing stable convergence and improved generalization.}
\label{fig:accuracy}
\end{figure}
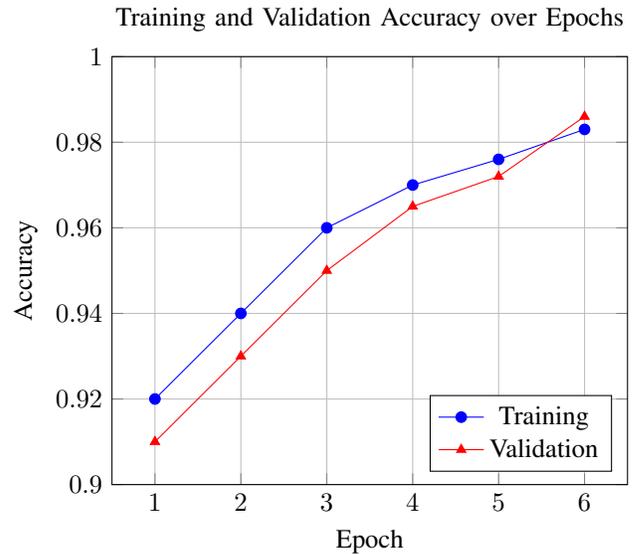

Figure~\ref{fig:accuracy} shows the training and validation accuracy curves. Both steadily increase and converge at high values with minimal gap, demonstrating good generalization and no overfitting. Regularization techniques such as dropout and early stopping contributed to this stability, and the smooth convergence indicates the model learned meaningful gait patterns across both datasets.

\begin{figure}[htbp]
\centering
\includegraphics[width=0.45\textwidth]{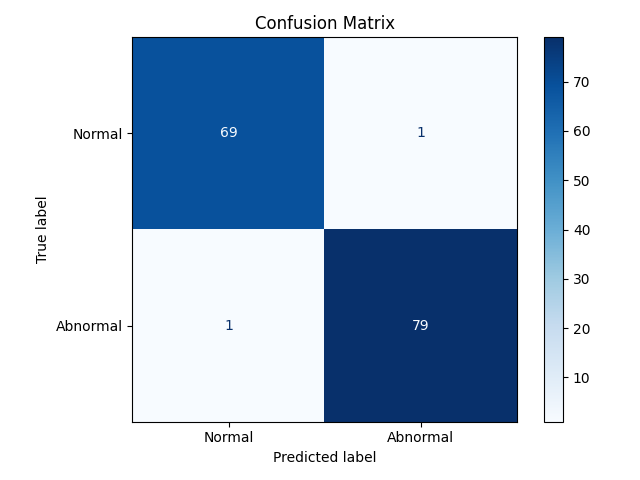}
\caption{Confusion matrix on the test set. The model demonstrates strong classification performance across both classes.}
\label{fig:confusion-matrix}
\end{figure}

Figure~\ref{fig:confusion-matrix} presents the confusion matrix for the GAVD test set. Predictions are strongly concentrated along the diagonal, showing accurate classification of both normal and abnormal gait. The small number of off-diagonal cases indicates high sensitivity and specificity, which is particularly important in clinical settings where missed abnormal detections carry high risk.

\begin{figure}[htbp]
    \centering
\includegraphics[width=0.5\textwidth]{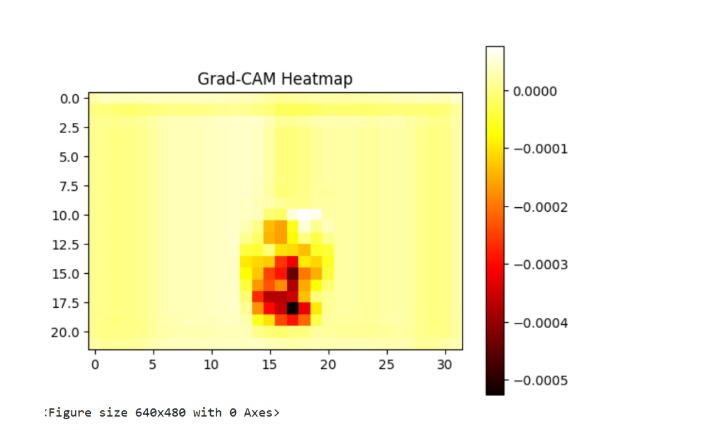}
    \caption{Grad-CAM heatmap visualization. This figure highlights the regions of the input most influential in the model's decision-making process.}
    \label{fig:gradcam-heatmap}
\end{figure}

On OU-MVLP, the model maintained strong performance across multiple viewing angles and subject identities. In Figure \ref{fig:gradcam-heatmap}, shows Grad-CAM heatmaps generated from OU-MVLP silhouettes. Regions in red and yellow highlight spatial areas most influential in classification decisions. These overlays reveal that the network focuses on clinically meaningful regions such as leg swing and posture during walking. Grad-CAM was applied specifically to OU-MVLP due to its visual input modality, providing interpretable insight into which spatial cues guided the model’s predictions.

By integrating Grad-CAM (spatial explanations on OU-MVLP) with SHAP (temporal explanations on GAVD), the framework provides complementary insights. Clinicians can see both where in the body the model focuses and when during the gait cycle features are most relevant. This dual perspective enhances transparency, supports trust in the predictions, and increases the likelihood of real-world adoption in both clinical diagnostics and large-scale biometric monitoring.

Overall, the model exhibits strong generalization, robust convergence, and dependable decision-making across both clinical and biometric datasets. Its consistent performance further supports its potential for real-world deployment in diagnostic or monitoring applications.

\subsection{Precision and Recall Dynamics}

To better understand the learning behavior of our model over time, we tracked the evolution of precision and recall across training epochs. These metrics are especially critical in clinical settings, where the costs of false positives and false negatives are not equivalent. \textbf{Recall} reflects the model’s sensitivity—its ability to correctly identify abnormal gait patterns—while \textbf{precision} reflects specificity, indicating how often predicted anomalies are actually correct.

\begin{figure}[htbp]
\centering
\includegraphics[width=0.9\linewidth]{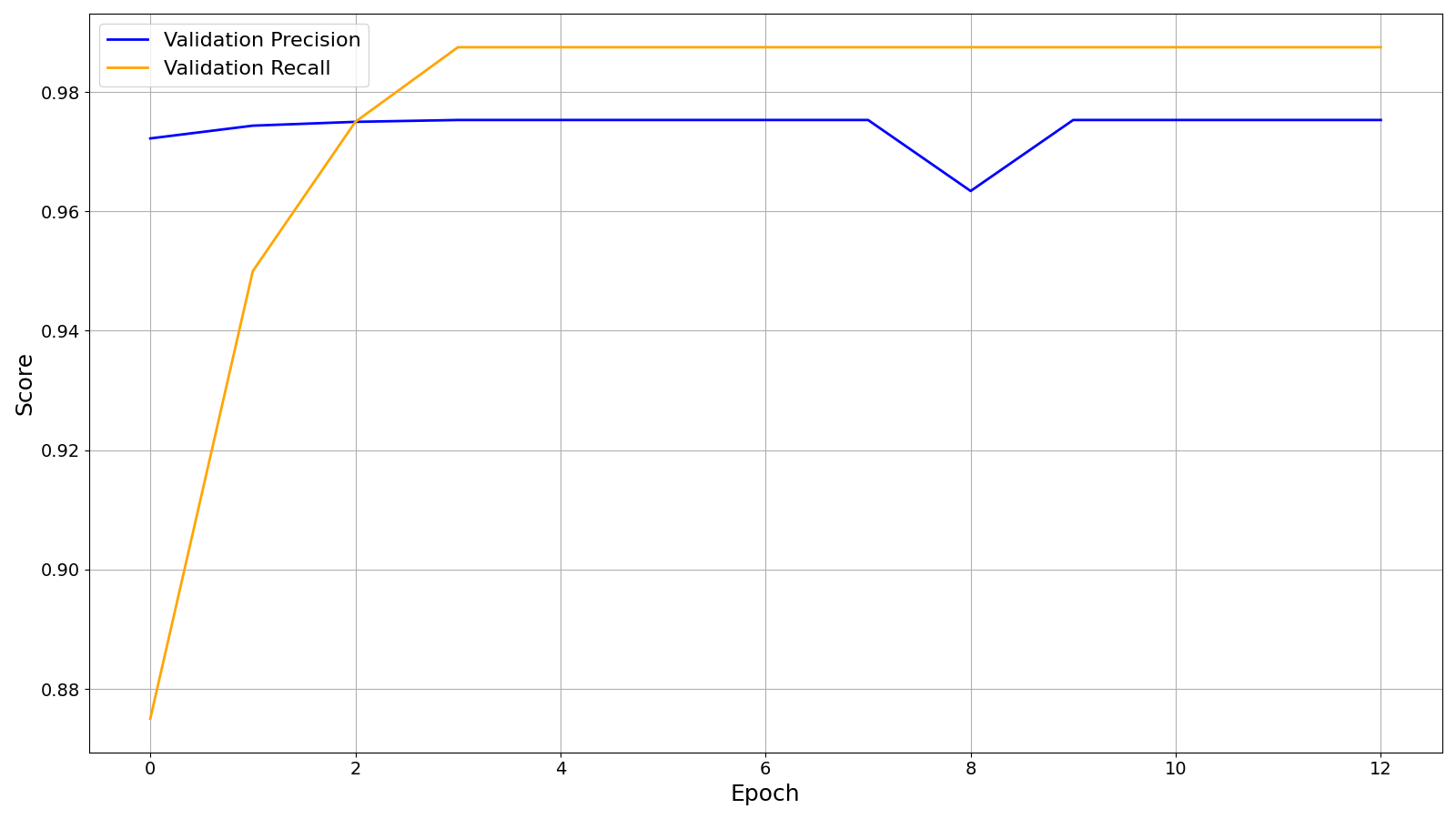}
\caption{Validation Precision and Recall across training epochs. Recall converges rapidly to 0.99, while precision remains high and stable with minor fluctuations, demonstrating the model’s ability to balance sensitivity and specificity in gait anomaly detection.}
\label{fig:precision-recall}
\end{figure}

Figure~\ref{fig:precision-recall}presents the validation precision and recall curves for the GAVD dataset. The recall curve rises steeply in the early epochs, rapidly reaching a stable value close to 0.988, indicating that the model is highly effective at identifying abnormal gait cases from the start. The precision curve, on the other hand, improves more gradually, stabilizing around 0.974–0.975, with minor fluctuations observed after epoch 4.

These small variations in precision can be attributed to the imbalanced class distribution and the presence of borderline gait sequences that are harder to consistently classify. This leads to a slight trade-off: while the model maintains high sensitivity (recall), it may occasionally misclassify normal gaits as abnormal, affecting precision.

Overall, the trends in Figure~\ref{fig:precision-recall} demonstrate that the model maintains a high recall without significantly compromising precision. 
In clinical or screening settings, this trade-off is acceptable, as favoring recall ensures that potential abnormal cases are not overlooked, and false positives can be handled through follow-up assessment. 
This balance between sensitivity and reliability contributes to the model’s practical utility in real-world gait analysis applications.

\subsection{Ablation Study}

To understand how each part of our model contributes to its overall performance, we ran an ablation study using the GAVD dataset. We tested how the model behaves when we remove or change key components: the LSTM layers, the second dataset (OU-MVLP), and the explainability tools (Grad-CAM and SHAP).

\begin{table}[htbp]
\caption{Ablation Study Results on the GAVD Dataset}
\centering
\begin{tabular}{|l|c|c|c|}
\hline
\textbf{Model Variant} & \textbf{Accuracy} & \textbf{Precision} & \textbf{F1-score} \\
\hline
CNN only (no LSTM) & 91.2\% & 0.88 & 0.89 \\
CNN + LSTM (GAVD only) & 96.8\% & 0.96 & 0.96 \\
CNN + LSTM + Explainability & 98.3\% & 0.98 & 0.98 \\
Full model (Dual dataset + XAI) & \textbf{98.6\%} & \textbf{0.99} & \textbf{0.99} \\
\hline
\end{tabular}
\label{tab:ablation}
\end{table}

Table~\ref{tab:ablation} shows the results. The first version used only CNN layers, without LSTM. This setup gave an accuracy of 91.2\%, and an F1-score of 0.89. 
These results suggest that spatial features alone are insufficient; including temporal information from the sequence is crucial for improved performance.
When we added LSTM layers to capture motion over time, accuracy improved to 96.8\%, and the F1-score rose to 0.96. This makes sense because walking patterns are inherently sequential, and LSTM helps capture that. Next, we included Grad-CAM and SHAP in the training pipeline. These tools not only make the model more interpretable but also seem to guide it toward better features. With them, performance improved further to 98.3\%. Finally, we trained the model using both GAVD and OU-MVLP datasets and kept the explainability tools. This full version achieved the best results: 98.6\% accuracy and an F1-score of 0.99. Combining different types of data helped the model generalize better, while the interpretability tools likely encouraged attention to more meaningful features.


In summary, each part of the model, such as temporal modeling, dataset diversity, and explainability, contributed valuable improvements. Together, they resulted in a more accurate and reliable gait classification system.

\section{Limitations and Future Work}

This study used two data sets, GAVD and OU-MVLP, which differ in format, size, and labeling. While this enabled testing across both clinical and biometric data, combining features from these sources proved challenging. Moreover, GAVD relies on pre-extracted joint data, complicating its real-time application in clinical settings without preprocessing.

One of the major limitations is the lack of direct clinical trials or patient-based validation. While the model’s predictions align with clinical observations and have been evaluated using clinical datasets, we did not conduct explicit trials with practitioners or patients to validate the system’s efficacy in real-world clinical settings. Future work will focus on conducting clinical trials to assess the model's real-world applicability and performance in collaboration with healthcare professionals. This will help validate the model’s utility and ensure it meets the necessary standards for clinical deployment.

Moving forward, we aim to enhance the model’s practical application by enabling it to process raw clinical videos without needing joint data extraction upfront. We are also exploring domain adaptation techniques to improve model performance across diverse datasets and clinical settings. Another focus will be to make the system more interpretable for healthcare professionals by providing visual feedback and involving them in the design process to ensure the model’s decisions are clear and actionable.

\section{Conclusion}

This paper presents a deep learning framework that not only detects gait abnormalities with high accuracy, but also explains its reasoning in ways that are useful to clinicians and practitioners. By combining CNN and LSTM, we capture both the posture and rhythm of walking, enabling the model to recognize abnormal patterns even under challenging conditions such as varied viewpoints or subtle timing shifts.

Our system outperforms conventional baselines and reaches 98.6\% accuracy on real-world gait datasets, including clinically labeled examples. Just as importantly, we make the model's decisions transparent using Grad-CAM and SHAP tools, which help pinpoint exactly which frames and which body regions signal an anomaly. This level of interpretability is rare in deep learning systems and essential for real-world trust and adoption.



\bibliographystyle{IEEEtran}
\bibliography{main}

\end{document}